\PassOptionsToPackage{table}{xcolor}
\documentclass[10pt,twocolumn,letterpaper]{article}

\usepackage[pagenumbers]{iccv} 

\usepackage{frcursive}
\usepackage{amsmath}
\usepackage{graphicx}

\definecolor{iccvblue}{rgb}{0.21,0.49,0.74}
\usepackage[pagebackref,breaklinks,colorlinks,allcolors=iccvblue]{hyperref}
\usepackage{amsfonts}
\usepackage{bbm}

\def\paperID{xxxx} 
\def\confName{ICCV}
\def\confYear{2025}

\usepackage{siunitx}
\usepackage{multirow}  
\usepackage{graphicx}  

\title{Seeing World Dynamics in a Nutshell}

\author{Qiuhong Shen\textsuperscript{1\footnotemark[1]},
        Xuanyu Yi\textsuperscript{2\footnotemark[1]},
        Mingbao Lin \textsuperscript{3},
        Hanwang Zhang \textsuperscript{2}, \\
        Shuicheng Yan \textsuperscript{3,1},
        Xinchao Wang \textsuperscript{1} \\
        \textsuperscript{1} National University of Singapore 
        \textsuperscript{2} Nanyang Technological University \\
        \textsuperscript{3} Skywork AI
}

\begin{document}

\twocolumn[{
\renewcommand\twocolumn[1][]{#1}
\maketitle
\centering
 \includegraphics[width=0.96\linewidth]{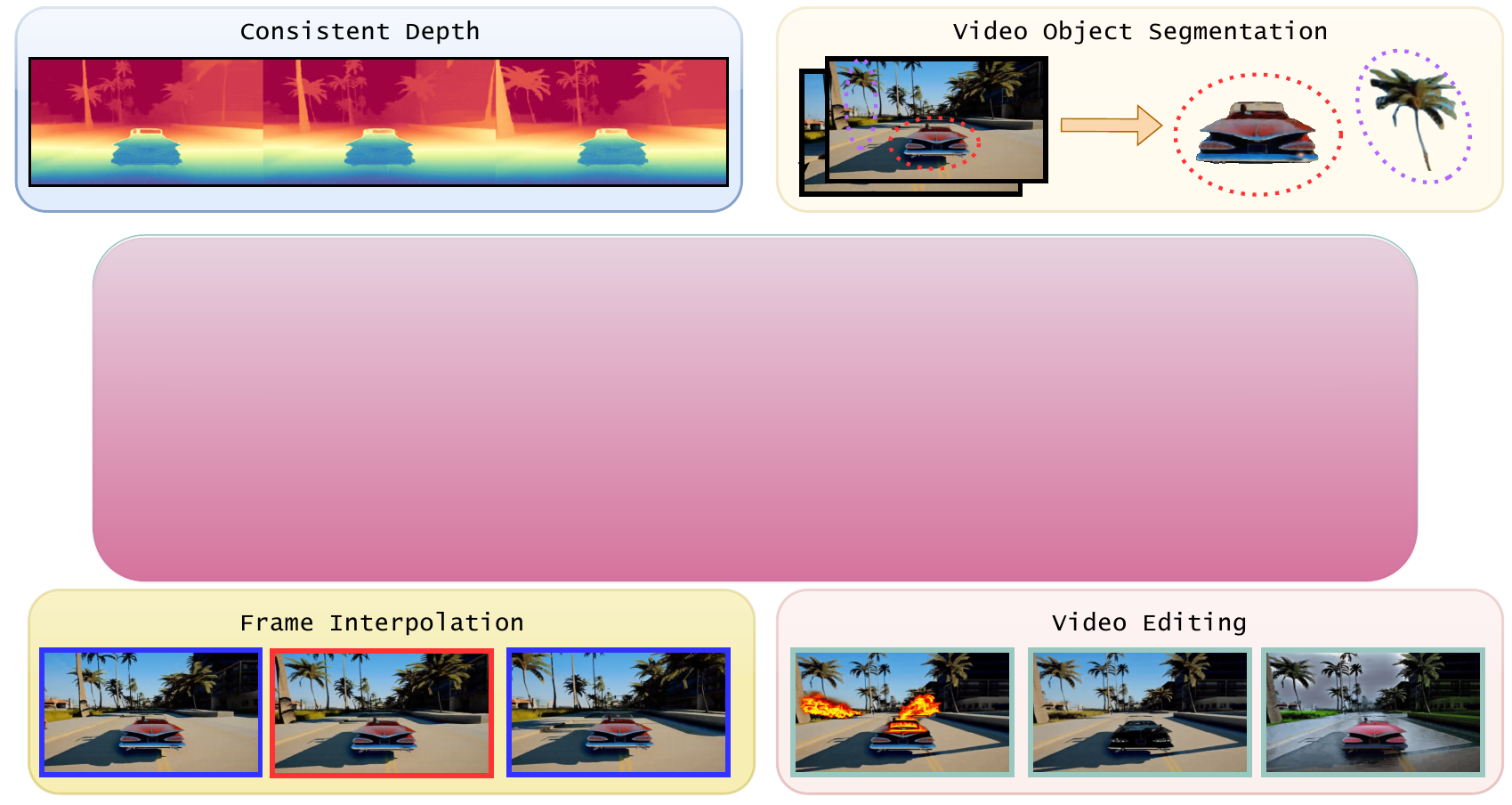}
 
 \captionsetup{type=figure}
 \vspace{-0.2cm}
\caption{We introduce NutWorld, a \textit{feed-forward} framework representing casual monocular videos parameterized by spatial-temporal aligned Gaussian (STAG), which empowers various video downstream processing tasks.}
\label{fig:intro}
}]

\maketitle

\renewcommand{\thefootnote}{\fnsymbol{footnote}}
\footnotetext[1]{Equal Contribution}
\footnotetext[3]{Work partially done in 2050 Research, Skywork AI}

\begin{abstract}
We consider the problem of efficiently representing casually captured monocular videos in a spatially- and temporally-coherent manner. 
While existing approaches predominantly rely on 2D/2.5D techniques treating videos as collections of spatiotemporal pixels, they struggle with complex motions, occlusions, and geometric consistency due to absence of temporal coherence and explicit 3D structure. 
Drawing inspiration from monocular video as a projection of the dynamic 3D world, we explore representing videos in their intrinsic 3D form through continuous flows of Gaussian primitives in space-time.
In this paper, we propose \textbf{NutWorld}, a novel framework that efficiently transforms monocular videos into dynamic 3D Gaussian representations in a single forward pass. 
At its core, NutWorld introduces a structured spatial-temporal aligned Gaussian (STAG) representation, enabling optimization-free scene modeling with effective depth and flow regularization. 
Through comprehensive experiments, we demonstrate that NutWorld achieves high-fidelity video reconstruction quality while enabling various downstream applications in real-time. 
Demos and code will be available at \href{https://github.com/Nut-World/NutWorld}{https://github.com/Nut-World/NutWorld}.

\end{abstract}

\vspace{-3mm}
\section{Introduction}

Our natural world exhibits inherent dynamism ----from rustling leaves swaying in the wind to clouds drifting across the sky and ocean waves rolling along the shore---where objects maintain their structural integrity while undergoing continuous spatiotemporal evolution. A fundamental objective in video processing~\cite{tekalp2015digital,bovik2009essential} is to enable machines to interpret such dynamic information, allowing recovery of both object geometry and motion patterns while preserving the spatiotemporal coherence intrinsic to human perception.
This capability is crucial for numerous applications, ranging from autonomous driving~\cite{gao2024vista,li2024vdg} and robotics~\cite{huang2023voxposer,huang2024rekep,o2023open} to augmented reality and content creation~\cite{sun2024sora,kondratyuk2023videopoet,bao2024vidu,yi2024diffusion}, where an accurate understanding of dynamic scenes directly impacts system performance and user experience.

Current neural-based video representations~\cite{ouyang2024codef, sitzmann2020implicit, ye2022deformable} predominantly rely on 2D and 2.5D techniques, treating videos as collections of spatiotemporal pixels~\cite{feichtenhofer2019slowfast, zhang2019comprehensive}. Although such discrete representations allow for basic temporal modeling through pixel matching~\cite{sun2018pwc} and tracking~\cite{karaev2023cotracker, ye2022ostrack}, they struggle to capture complex scene dynamics and maintain appearance consistency, particularly in scenarios involving occlusions and non-rigid deformations~\cite{newcombe2015dynamicfusion, ouyang2024codef}. Moreover, they inherently lack explicit 3D understanding, leading to unreliable motion captures and spatial arrangements.

Drawing inspiration from the fact that a monocular video is a projection of the dynamic 3D world, we ask: \textit{Can videos be represented in a canonical 3D space form without per-scene optimization}? Recent advances in dynamic Gaussian Splatting~\cite{kerbl20233d} have shown remarkable capabilities in dynamic scene reconstruction~\cite{yang2024deformable, wu20244d, duan20244d} and video Gaussian representation~\cite{sun2024splatter, wang2024gflow, lei2024mosca}, achieving high-fidelity rendering with explicit 3D representations, albeit through per-scene optimization.
By modeling videos as flows of Gaussian primitives over time, we overcome the limitations of 2D representations and enable video Gaussian representation without per-scene optimization. As illustrated in Figure~\ref{fig:intro}, this paradigm treats space-time holistically and offers key advantages: each Gaussian acts as a flexible building block that adapts to local appearance structures, enabling seamless representation of complex scenes. Moreover, when endowed with dynamic attributes, these structured Gaussians naturally approximate the underlying spatiotemporal volume of monocular videos, capturing intrinsic motions with temporal consistency and facilitating various downstream tasks such as object segmentation~\cite{flashsplat, sun2024splatter}, video editing~\cite{ceylan2023pix2video, qi2023fatezero}, and frame interpolation~\cite{dong2023video, lu2022video}.


%

However, transforming casually captured monocular videos to dynamic Gaussian representations \textit{instantly (in seconds per frame)} poses several challenges: 
\textbf{ (1) Unposed inputs}. Gaussian Splatting~\cite{kerbl20233d} and its variants~\cite{yu2024mip, yu2024gaussian, huang20242d} heavily rely on accurate camera poses obtained through Structure-from-Motion (SfM)~\cite{ullman1979interpretation, schonberger2016structure}, which are typically unavailable for casual monocular videos. Without such pose guidance, current methods~\cite{Fu_2024_CVPR, bian2023nope, lin2021barf, wang2021nerf} fail to disentangle camera motion from scene dynamics, resulting in deteriorated rendering quality and even collapsed reconstruction. 
\textbf{(2) Nonstructural Nature}. Most existing dynamic Gaussian Splatting either leverage per-scene optimized deformation networks~\cite{yang2024deformable, wu20244d, duan20244d, katsumata2025compact, kratimenos2025dynmf, yang2023real} or adopt per-frame Gaussian generation~\cite{ren2024l4gm} for dynamics modeling, both incompatible with our feed-forward prediction paradigm. 
The spatially unstructured property of Gaussian primitives further makes them prone to local minima in inverse rendering~\cite{zhu2025fsgs, chung2024depth}, impeding feed-forward modeling of dynamic underlying structures in monocular videos.
\textbf{(3) Spatial Ambiguity}. The absence of multi-view supervision and initialization from SfM points significantly limits the spatial modeling capability of Gaussian Splatting, leading to ambiguous scale, depth collapse, and inconsistent spatial arrangements in reconstructed scenes.



To address these challenges, we introduce \textit{NutWorld} to efficiently transform monocular videos into dynamic Gaussian Splatting representations in a \textit{single forward pass}. Our method has three key components: (1) A structured spatial-temporal aligned Gaussian (STAG) representation in a canonical space (Section~\ref{stag}), enabling feed-forward prediction with pose-free, scale-invariant modeling. (2) An feed-forward pipeline (Section~\ref{NutShell}) that learns spatial-temporal correspondences and  motions across frames, swiftly transforming them into STAG representations. (3) Depth and flow regularization (Section~\ref{prior}), leveraging calibrated depth~\cite{chen2025video, yang2024depth} and optical flow priors~\cite{xu2023unifying} to resolve spatial ambiguity and motion-appearance entanglement in the ill-posed monocular setting~\cite{sun2024splatter, som2024}. With large-scale pre-training, \textit{NutWorld} processes arbitrarily long videos while preserving spatial-temporal consistency through segment-based inference (Section~\ref{train}).


%

%

We performed qualitative and quantitative experiments on RealEstate10K~\cite{zhou2018stereo} and MiraData~\cite{ju2024miradatalargescalevideodataset} to verify the efficacy of \textit{NutWorld} in video reconstruction. Moreover, our method demonstrates real-time inference speed and flexibility across various downstream tasks, including novel view synthesis, consistent depth estimation, video segmentation, video editing, and frame interpolation, indicating its potential as a general-purpose video representation framework. Our main contributions include:
\begin{itemize}[leftmargin=*]
\item We present the \textit{first} framework to efficiently represent world dynamics in casually captured monocular videos via dynamic Gaussian Splatting in a \textit{single forward pass}.

\item Our NutWorld framework incorporates the STAG representation, elaborated network for feed-forward reconstruction, and effective regularization strategies for spatial and temporal coherent recovery from casual videos.

\item Extensive experiments on video reconstruction and various downstream tasks confirm the spatial-temporal coherence and versatility of NutWorld.
\end{itemize}








\label{sec:intro}

\section{Related Work}

\noindent\textbf{Neural video representation.}  
Casually captured monocular videos are widely available and can be considered as 2D projections of dynamic 3D scenes. Efficient representations of these videos are crucial for various computer vision tasks, including video object segmentation, object tracking, depth estimation, and frame interpolation.
Early works~\cite{chen2021learning, sitzmann2020implicit, tancik2020fourier, ouyang2024codef} leveraged implicit neural representations, modeling images through coordinate-based Multilayer Perceptrons (MLPs) through per-image optimization. Later works expanded upon this by incorporating learnable deformation field~\cite{ouyang2024codef, wang2023tracking, liu2023robust, ye2022deformable}, reconstructing videos as canonical and deformable MLPs. However, these methods often struggle to capture complex motions due to the absence of explicit 3D structure. Recently, explicit Gaussian video representations~\cite{som2024, sun2024splatter, wang2024gflow, lei2024mosca} have emerged to address these limitations, representing videos explicitly as Gaussians in a 3D canonical space, each associated with a 3D motion projected onto the 2D video frames. But these methods still require computationally expensive per-video optimization, limiting their practical application.
In contrast, our NutWorld introduces a novel feed-forward Gaussian video representation, distinguishing itself from previous approaches. The NutWorld network, trained on video datasets, efficiently reconstructs videos as Gaussians through a single forward pass, delivering improved reconstruction quality with substantial speedup.

\noindent\textbf{Feed-Forward Gaussian Splatting.} Recent advance in large-scale 3D scene datasets~\cite{ling2024dl3dv, zhou2018stereo, liu2021infinite} has enabled feed-forward Gaussian approaches~\cite{pixelsplat, chen2025mvsplat, zhang2024transplat, zhang2025gs, wewer2024latentsplat,yi2024mvgamba,shen2024gamba}, which excel in efficiency and sparse view reconstruction. PixelSplat~\cite{pixelsplat} and LatentSplat~\cite{wewer2024latentsplat} employ the epipolar transformer~\cite{he2020epipolar, wang2022mvster} to establish cross-view correspondences and predict 3D Gaussians from multi-view image features, while MVSplat~\cite{chen2025mvsplat} utilizes cost volumes to jointly predict depth and Gaussian parameters. Alternatively, GS-LRM~\cite{zhang2025gs} and Long-LRM~\cite{ziwen2024long} consider Gaussian Splatting reconstruction as a sequence-to-sequence translation task, employing transformer-based~\cite{vaswani2017attention} or hybrid~\cite{mamba2} architectures to regress Gaussian primitives. 
However, these feed-forward reconstruction methods designed for static 3D scenes have limitations when generalized to unconstrained videos, primarily because accurate per-frame camera poses cannot be obtained even through advanced pose estimation methods~\cite{wang2024vggsfm, wang2024dust3r, zhang2024monst3r, mast3r}. To address this, we introduce a canonical camera space for building our NutWorld model, enabling robust 3D representation of dynamic scenes.

\label{sec:formatting}

\vspace{-2mm}
\section{Preliminary: Dynamic Gaussian Splatting}
Dynamic Gaussian Splatting~\cite{yang2024deformable, wu20244d, duan20244d, katsumata2025compact, kratimenos2025dynmf, yang2023real} is an explicit 4D neural representation for reconstructing dynamic 3D scenes from multi-view videos through differentiable rasterization~\cite{kerbl20233d}. Dynamic Gaussian ${\{\mathcal{G}_{i}, \mathcal{D}_{i}(t)}\}$ decouples dynamic scenes into a static canonical 3D Gaussian $\mathcal{G}_i$ and a deformation motion field $\mathcal{D}_i$ to account for temporal variations in 3D space. Specifically, the static 3D Gaussian $\mathcal{G}_{i}$ is composed of a 3D center $\mu \in \mathbb{R}^3$,  3D scale $s \in \mathbb{R}^3$, associated color $c \in \mathbb{R}^3$, opacity $\alpha \in \mathbb{R}$, and rotation quaternion $q \in \mathbb{R}^4$. 
%
%
For the deformation fields $\mathcal{D}_{i}(t) = \{\mu_i(t), q_i(t), \alpha_i(t)\}$, the deformable attributes and their parameterizations vary across different approaches but are generally limited to the center position $\mu_i$, rotation $q_i$, and opacity $\alpha_i$, with other attributes remaining independent of time.
When representing the scene at time $t$, each dynamic Gaussian is temporally sliced into 3D space by applying its corresponding deformation field $\mathcal{D}_{i}(t)$ to yield an static Gaussian primitive, e.g., with deformed position $\hat{\mu}_i = \mu_i + \mu_i(t)$, which can then be efficiently rendered into 2D images via tile-based rasterization pipeline.



\vspace{-1.5mm}

\section{Methodology}
\begin{figure*}[t]
\centering 
\includegraphics[width=0.97\linewidth]{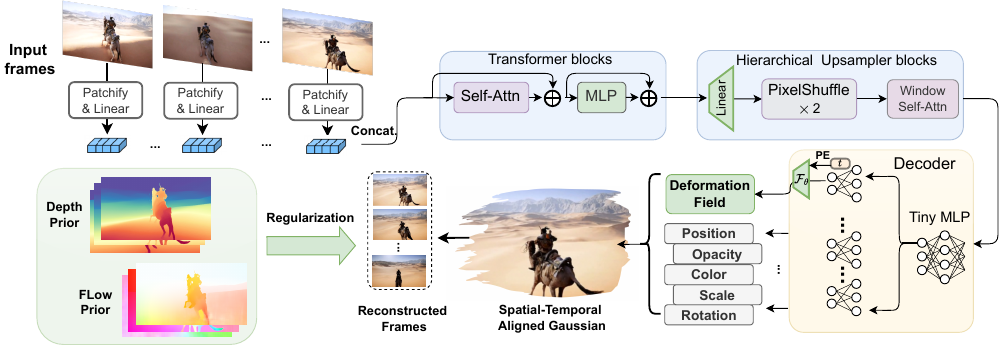} 
\vspace{-10pt}
\caption{\textbf{Overview of NutWorld.} We directly predict STAG in a canonical space from sparse input frames via a transformer-based reconstruction model, where calibrated depth and flow priors are leveraged to avoid depth ambiguity and motion uncertainty.}
\label{fig:framework}
\vspace{-1mm}
\end{figure*}

In this section, we present a framework for efficiently representing world dynamics from monocular video in a \textit{feed-forward manner}. As shown in Figure~\ref{fig:framework}, we first introduce our Spatial-Temporal Aligned Gaussian Splatting (STAG) representation (Section~\ref{stag}). To enable the mapping of videos to STAG in a single forward pass, we detail our transformer-based network (Section~\ref{NutShell}), which operates with calibrated depth and flow priors (Section~\ref{prior}). Finally, we discuss the overall training objectives and protocols for processing long video segments (Section~\ref{train}).



\begin{figure}[t]
\centering
\vspace{-5mm}
\includegraphics[width=1.0\columnwidth]{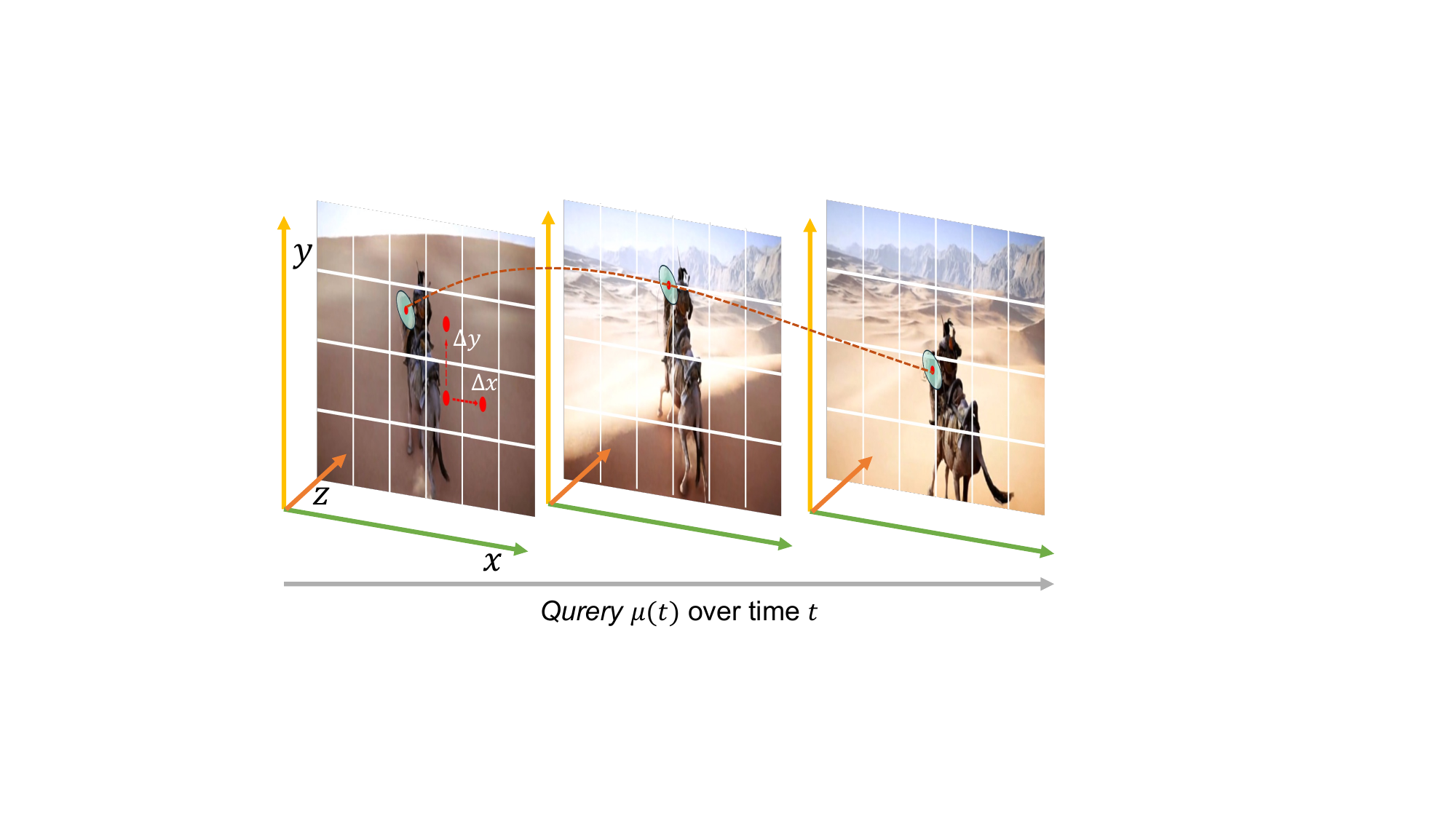}
\caption{The illustration of STAG to represent dynamic scenes.}
\label{fig:stag}
\end{figure}

\subsection{Spatial-Temporal Aligned Gaussian}
\label{stag}
\noindent \textbf{Canonical camera space}. Given an unposed monocular video, we employ an orthographic camera coordinate system to interpret the input as a quasi-3D canonical volume~\cite{wang2023tracking,sun2024splatter} rather than an absolute 3D world space. This choice addresses two challenges: (1) the difficulty of obtaining consistent camera trajectories in dynamic scenes~\cite{ullman1979interpretation, schonberger2016structure, pan2024global, wang2024dust3r, wang2024vggsfm}, and (2) the scale ambiguity in feed-forward 3D reconstruction~\cite{pixelsplat, chen2025mvsplat, zhang2024transplat}, where perspective projection couples object size with camera distance. By imposing a fixed pose along the $z$ axis, orthographic projection removes perspective-induced distortions and eliminates the need for explicit camera estimation, enabling joint modeling of both camera and object motion. We detail the orthographic rasterization pipeline in the \textit{Appendix}.

\noindent \textbf{Structured Dynamic Gaussian}. To overcome the unstructured nature in dynamic Gaussian Splatting and facilitate neural network integration, we introduce \textbf{S}patial-\textbf{T}emporal \textbf{A}ligned \textbf{G}aussian Splatting (\textbf{STAG}) within this canonical volume. STAG constrains each dynamic Gaussian to a specific pixel location and timestamp, in contrast to the previous approach of predicting unconstrained Gaussians with deformable fields across orthogonal space-time.
Formally, for an input frame $F_{k}$ with normalized timestamp $t_k \in [0, 1]$, we compute a Gaussian feature map $\mathcal{E}^k \in \mathbb{R}^{U \times V \times T}$, where $U$ and $V$ represent spatial dimensions, and $T$ denotes the channel dimension. Each $T$-dimensional pixel is decoded into a 3D Gaussian with an associated deformation field $\{\mathcal{G}_{i}, \mu_i(t)\}$~\footnote{ Note that based on our empirical observation, we simplify the deformation modeling by considering only the center position, such that the deformation field reduces to $\mathcal{D}_{i}(t) = \mu_i(t)$.} in a pixel-aligned manner~\cite{pixelsplat, szymanowicz2024splatter}.

Given a pixel at coordinates $(u, v)$ in the feature map $\mathcal{E}^k$, we define its corresponding 3D Gaussian center $\mu^k$ through unprojection along the corresponding ray: $\mu^k = (u + \Delta_{x}, v + \Delta_{y}, d)$,
where $\Delta_{x}$ and $\Delta_{y}$ are bounded offsets decoded from $\mathcal{E}^k$, maintaining pixel-level correspondence with local position refinement. The depth value $d$ specifies the $z$-coordinate of $\mu^k$ along the camera's viewing axis. To model temporal dynamics, for frame rendering at timestamp $t_j$, we deform each static Gaussian center $\mu^k$ during frame rendering at timestamp $t_j$ using the predicted deformation field $\mu^k(t)$:
\begin{equation}
\hat{\mu}^k = \mu^k + \mu^k(t_{j}) \mathbbm{1}(t_k, t_j),
\end{equation}
where $\mathbbm{1}(t_k, t_j)$ is a temporal slicing indicator function defined as:
\begin{equation}
\mathbbm{1}(t_k, t_j) = \left\{\begin{array}{ll}
0, & \text{if} \,\, t_j = t_k \text{ \small{\textcolor{gray}{(reference frame)}}}, \\
1, & \text{otherwise \text{\small{\textcolor{gray}{(non-reference frame)}}}}. 
\end{array}\right.
\label{eq:temporal_align}
\end{equation}
As shown in Figure~\ref{fig:stag}, the temporal slicing function $\mathbbm{1}(t_k, t_j)$ modulates the deformation field $\mu^k(t_{j})$ based on the temporal relationship between the rendering timestamp $t_j$ and the Gaussian's reference timestamp $t_k$. For $t_j = t_k$, the deformation is suppressed ($\mathbbm{1}(t_k, t_j) = 0$), preserving the original Gaussian position $\mu^k$ to maintain spatial alignment. When $t_j \neq t_k$, the deformation field is activated to adapt $\mu^k$ according to temporal differences, enabling per-pixel alignment across frames and enhancing consistency in our quasi-3D volume.


\subsection{Encapsulate Dynamics within ``Nutshell''}
\label{NutShell}
In this section, we introduce the transformer-based model in NutWorld to transform unposed monocular videos into the proposed STAG. Formally, given an input sequence of $K$ video frames $\{F_{k}, t_{k} \}$, where each frame $F_{k} \in \mathbb{R}^{H \times W \times 3}$ is associated with a normalized timestamp $t_{k} \in [0, 1]$, we define NutWorld as an inverse mapping function $\Theta$:
\begin{equation}
\{\mathcal{G}_{i}, \mu_{i}(t)\} = \Theta(\{F_{k}, t_{k}\}).
\label{eq:model}
\end{equation}
This mapping generates a set of STAGs $\{\mathcal{G}_{i}, \mu_{i}(t)\}$ that can be rendered into arbitrary frames ($M \geq K$) through temporal interpolation of the deformation field. Specifically, the NutWorld model is composed of two main components:


\noindent{\textbf{Transformer-based Encoder}}. For each input frame $F_{k} \in \mathbb{R}^{H \times W \times 3}$, we augment pixels by concatenating their RGB values with corresponding depth coordinates $d^{*}_{k}$ along the channel dimension, obtained from a pre-trained video depth estimation model~\cite{chen2025video}. The augmented frames are first split into non-overlapping patches of size $p \times p$ using convolutions, which are then linearly transformed and concatenated across all frames to generate transformer input tokens. Notably, our architecture eliminates the need for explicit positional embeddings as used in ViT~\cite{dosovitskiy2020image,bai2024meissonic}, since depth coordinates inherently encode spatial information. The transformer blocks, comprising self-attention~\cite{vaswani2017attention} and MLP layers, process these concatenated tokens to capture spatiotemporal correspondence and produce encoded features $\mathcal{E}_{0} \in \mathbb{R}^{K \times h \times w \times C}$, where $h = H/p$ and $w = W/p$ denote the spatial resolution and $C$ denotes the feature dimension. To ensure sufficient STAGs for casual videos, we leverage a hierarchical upsampling network~\cite{zhang2025gs, xu2024grm} that progressively expands the spatial resolution of the encoded feature $\mathcal{E}_0$.  Each upsampler block first expands the feature dimension by a factor of \textbf{\textit{4}} through a linear layer, followed by a PixelShuffle~\cite{shi2016real} layer that doubles the spatial resolution. The resulting features are then processed by a local attention layer with window size $\mathcal{W}$, which balances computational efficiency and spatial-temporal feature aggregation:
\begin{equation}
\begin{aligned}
\hat{\mathcal{E}}_{j-1} &= \text{PixelShuffle}(\text{Linear}(\mathcal{E}_{j-1}), 2), \\
\mathcal{E}_{j} &= \text{WindowAttn}(\hat{\mathcal{E}}_{j-1}, \mathcal{W}).
\end{aligned}
\end{equation}
After cascading such blocks $n$ times, the final feature map $\mathcal{E}_{n}$ achieves a spatial resolution of $(U, V) = (2^n h, 2^n w)$.

\noindent{\textbf{STAG Decoder}}. The proposed decoder predicts both static Gaussian attributes and their deformable field from the upsampled feature map $\mathcal{E}_{n}$.
For decoding static 3D Gaussian, we employ a shared MLP with specialized sub-heads to predict each Gaussian attribute: center position $\mu$, opacity $\alpha$, scale $s$, rotation $q$, and color $c$ is RGB value $\in [-1, 1]^3$. Given our fixed canonical camera setup, we omit view-dependent effects in color prediction. Each attribute utilizes specific activation functions following established practices~\cite{pixelsplat}, with details provided in the \textit{Appendix}.

To model the deformation field $\mu(t)$ as a continuous function of time, we encode the timestamp $t$ using sinusoidal positional encoding followed by a learnable linear projection. The encoded time features are then processed by an MLP to enable differentiable temporal interpolation:
\begin{equation}
\mu(t) = \mathcal{F}_{\theta}(\text{Linear}(\gamma_{n}(t)), \mathcal{E}_n(k, u, v)),
\end{equation}
where $\mathcal{F}_{\theta}$ represents an MLP with learnable parameters $\theta$. A $tanh$ activation is applied to the output, constraining the deformation within the bounds $[-b, b]^3$. The function $\gamma(t) = \left( \sin(2^k \pi t), \cos(2^k \pi t) \right)_{k=0}^{L-1}$ represents the sinusoidal expansion of order $L$, which enhances the network's capacity to capture high-frequency components effectively.


\subsection{Calibrated 2D Priors Regularization}
\label{prior}


Learning spatially-aware STAGs solely from monocular videos is inherently ill-posed due to depth ambiguity and motion uncertainty. Therefore, we leverage off-the-shelf foundation models~\cite{xu2023unifying, chen2025video} to recover spatial relationships and temporal motion through calibrated depth and flow priors, respectively. Note that in the quasi-3D canonical volume under the orthographic projection, the movement of STAG along the xy-coordinates directly corresponds to optical flow magnitude, whereas depth-related loss exclusively affects the z-coordinate, further facilitating the incorporation of those 2D priors. 

\noindent\textbf{Depth Regularization}. To enhance robustness against scale and shift variations in depth rendering, we employ a scale and shift invariant loss~\cite{bhat2023zoedepth, ranftl2020towards}. This loss function computes optimal scaling $\beta$ and shifting $\gamma$ factors that align the rendered depth $d$ with the pseudo depth $d^*$ estimated by the video depth prior~\cite{chen2025video}. 
The optimal values for $\beta$ and $\gamma$ are obtained by minimizing the squared error between the scaled rendered depth and the pseudo depth, as follows:
\begin{equation}
\begin{aligned}
    \beta, t &= \arg \min_{\beta, \gamma} \sum_{i} M_{i} \left( \beta \cdot d_{i} + \gamma - d^{*}_{i} \right)^2, \\
    \mathcal{L}_{\text{depth}} &= \sum_{i=1}^{H \times W} | \tilde{d}_{i} - d^{*}_{i} | / {\sum M}, \quad \tilde{d}_{i} = \beta \cdot d_{i} + \gamma.
\end{aligned}
\label{eq:depth_prior}
\end{equation}
Here, $M$ is an outlier mask where $M_i=0$ for the top $10\%$ values in each depth map and $M_i=1$ otherwise, mitigating noise in the estimated pseudo depth $d^*$. This depth supervision effectively regularizes the training, making it robust for predicting relative depth in scenes.

\begin{figure*}[t]
\centering 
\includegraphics[width=0.9\linewidth]{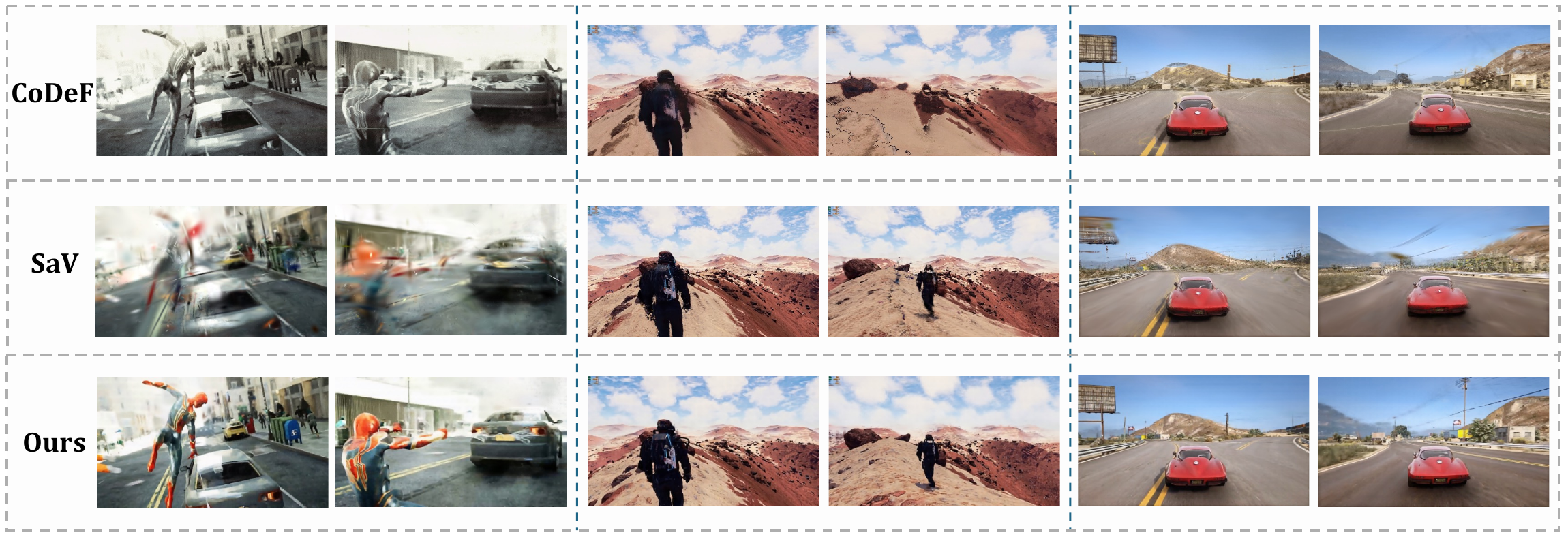} 
\caption{Qualitative comparison of video reconstruction using our NutWorld and other optimization-based methods.}
\label{fig:main-comparison}
\end{figure*}

\begin{table*}[htbp]
  \centering
  \setlength{\tabcolsep}{6pt} 
  \renewcommand{\arraystretch}{1.0}  
  \resizebox{0.85\linewidth}{!}{\begin{tabular}{l S[table-format=2.2] S[table-format=1.4] S[table-format=1.4] l S[table-format=3.1] l}
    \toprule[1pt]  
    \rowcolor[HTML]{EFEFEF}
    \textbf{Method} & {\textbf{PSNR}$\uparrow$} & {\textbf{SSIM}$\uparrow$} & {\textbf{LPIPS}$\downarrow$} & {\textbf{Inference Time}} & {\textbf{GPU Mem.}} & {\textbf{FPS}} \\
    \midrule[0.6pt]  
    4DGS~\cite{wu20244d} & {17.59} & {0.5725} & {0.5393} & {\small$\sim$40 mins} & {10G} & {145} \\
    RoDynRF~\cite{liu2023robust} & {24.63} & {0.7152} & {0.3927} & {\small$>$24 hours} & {24G} & {$<$ 0.01} \\
    CoDeF~\cite{ouyang2024codef} & {26.35} & {0.8045} & {0.2714} & {\small$\sim$30 mins} & {10G} & {8.8} \\
    Splatter-a-Video~\cite{sun2024splatter} & {24.85} & {0.7413} & {0.3523} & {\small$\sim$30 mins} & {10G} & {149} \\
    \midrule[0.4pt]  
    \textbf{Ours} & {29.18} & {0.9015} & {0.1415} & {\small$1.8$ seconds} & {4G} & {450} \\
    \bottomrule[1pt]  
  \end{tabular}}
  \caption{Quantitative comparison with state-of-the-art methods. GPU Memory is measured in GB and FPS indicates rendering frame rate.}
  \vspace{-6mm}
  \label{tab:main-comparison}
\end{table*}

\noindent\textbf{Flow Regularization}.  We extract global STAG trajectories by leveraging frame-to-frame optical flow associations~\cite{xu2023unifying}. In contrast to previous methods that solely employ iterative optimization between adjacent frames, our feed-forward framework utilizes global trajectory supervision to ensure consistent motion in a single forward pass. 

For each STAG, we define its estimated pseudo-trajectory $\bar{\mu}^*(t)$ across $K$ video frames. This trajectory is derived by sequential queries to the pre-computed optical flow field between adjacent frames, represented by a global flow matrix \(\mathbf{F}\):
\vspace{-2mm}
\begin{equation}
\vspace{-1mm}
\resizebox{0.95\columnwidth}{!}{$
\mathbf{F} = \begin{pmatrix}
(0, 0, \dots, 0) & \mathbf{f}_{1 \to 0} & \mathbf{f}_{2 \to 1} + \mathbf{f}_{1 \to 0} & \dots & \sum\limits_{k=1}^{K-1} \mathbf{f}_{k \to k-1} \\
\mathbf{f}_{0 \to 1} & (0, 0, \dots, 0) & \mathbf{f}_{2 \to 1} & \dots & \sum\limits_{k=2}^{K-1} \mathbf{f}_{k \to k-1} \\
\mathbf{f}_{0 \to 1} + \mathbf{f}_{1 \to 2} & \mathbf{f}_{1 \to 2} & (0, 0, \dots, 0) & \dots & \sum\limits_{k=3}^{K-1} \mathbf{f}_{k \to k-1} \\
\vdots & \vdots & \vdots & \ddots & \vdots \\
\sum\limits_{k=1}^{K-1} \mathbf{f}_{k-1 \to k} & \sum\limits_{k=2}^{K-1} \mathbf{f}_{k-1 \to k} & \sum\limits_{k=3}^{K-1} \mathbf{f}_{k-1 \to k} & \dots & (0, 0, \dots, 0)
\end{pmatrix}
$}
\end{equation}
The global flow matrix $\mathbf{F}$ is structured as a $K \times K$ matrix, where each entry $\mathbf{F}_{i, j}$ is structured as a vector of length $(U \times V)$ represents the 2D cumulative motion of each Gaussian from frame $j$ to frame $i$. The matrix structure is asymmetric: upper triangular entries encode cumulative backward flow $\mathbf{f}_{k \to k-1}(\bar{\mu}_{i})$, while lower triangular entries contain cumulative forward flow $\mathbf{f}_{k \to k+1}(\bar{\mu}_{i})$. Here, $\bar{\mu}_{i}$ denotes the projected 2D coordinates of the $i$-th Gaussian in frame $k$. For notational clarity, we omit the explicit flow query operations in the matrix entries.

Using the global flow matrix, we regularize the deformation field $\mu(t)$ by comparing its 2D projection $\bar{\mu}(t)$ against the estimated global trajectories $\mu^*(t)$ across $K$ frames:
\begin{equation}
\mathcal{L}_{\text{flow}} = \sum_{i=1}^{K \times U \times V}\sum_{k=0}^{K-1} \|\bar{\mu}_{i}(t_k) - \mu^*_i(t_k)\|_1,
\label{eq:flow_prior}
\end{equation}
where $\|\cdot\|_1$ denotes the L1 norm. An outlier filtering strategy is also applied to calibrate the flow prior (see \textit{Appendix} for flow calibration details). This flow loss enforces consistency between the predicted trajectories and the reference paths derived from optical flow, enabling NutWorld to learn coherent motion patterns from casually captured videos.




\subsection{Training and Inference}
\label{train}
\noindent\textbf{Overall objective.} During the training phase, we render RGB frames from $K = 6$ sparsely sampled frames and interpolate $M = 10$ intermediate frames for dense temporal supervision. Our training objective comprises three terms:

\begin{equation}
\mathcal{L} = \mathcal{L}_{\text{MSE}} + \lambda_{\text{flow}}\mathcal{L}_{\text{flow}} + \lambda_{\text{depth}}\mathcal{L}_{\text{depth}},
\end{equation}
where $\mathcal{L}_{\text{MSE}}$ is the mean square error loss between the rendered and ground truth RGB frames. $\mathcal{L}_{\text{flow}}$ and $\mathcal{L}_{\text{depth}}$ denote the calibrated regularization term, respectively. The coefficients $\lambda_{\text{flow}}$ and $\lambda_{\text{depth}}$ balance the contribution of each term.

\noindent\textbf{Segment-based long video inference.} To handle casual videos with hundreds of frames, we propose a simple but effective segment-based strategy during inference. The input video is divided into overlapping segments, where the adjacent segments share one frame. Due to our pixel-level spatial-temporal representation, Gaussian trajectories can be seamlessly propagated across segments through these shared frames, enabling NutWorld to process arbitrarily long videos while maintaining spatial-temporal consistency. The details of segment-based inference is in \textit{Appendix}.




\section{Experiment}
\label{sec:exp}



\begin{figure*}[t]
\setlength{\abovecaptionskip}{0.2cm}
\setlength{\belowcaptionskip}{0cm}
\centering 
\includegraphics[width=0.9\linewidth]{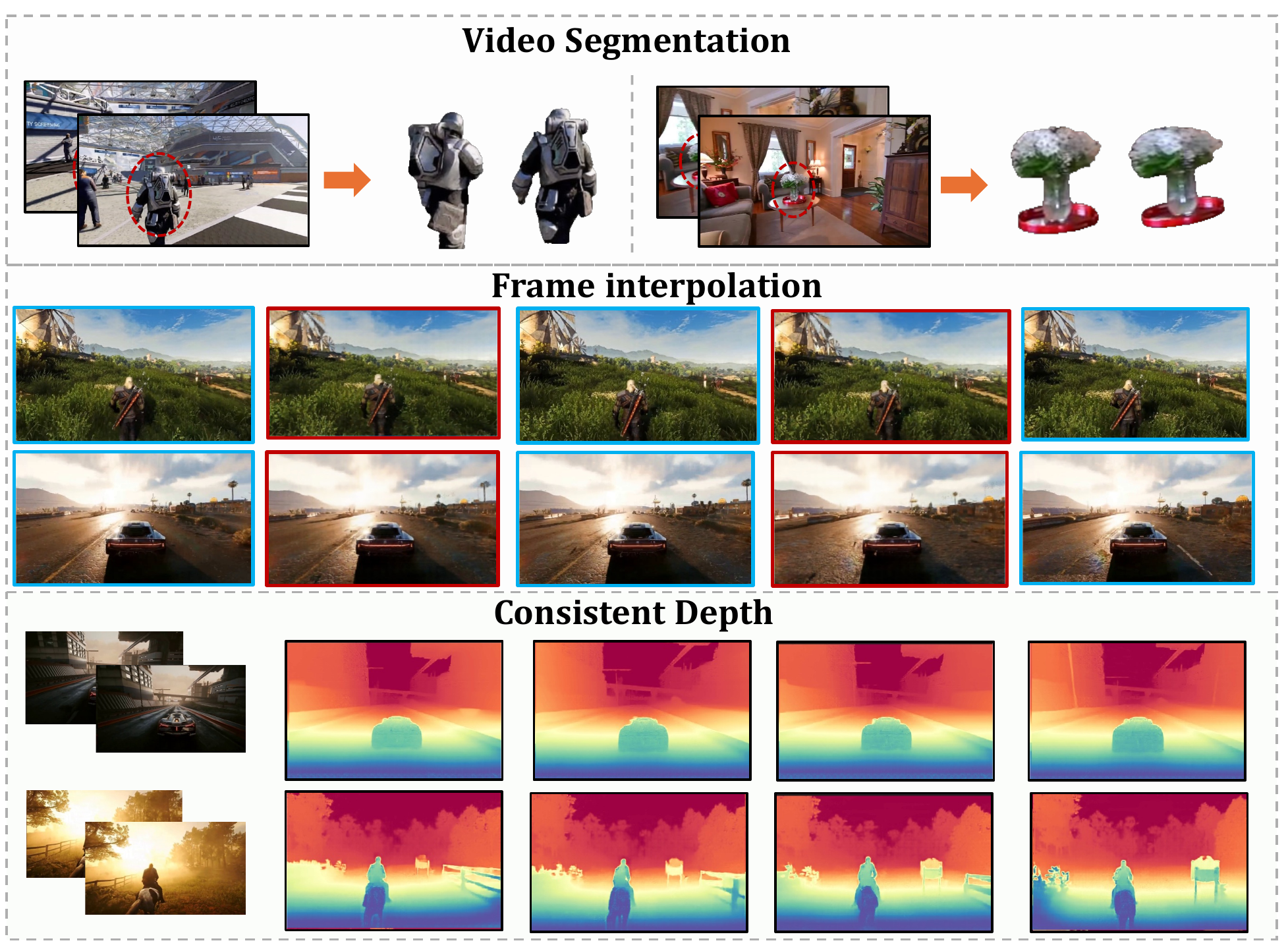} 
\vspace{-8pt}
\caption{Qualitative results in various downstream tasks, including video segmentation, editing, frame interpolation and consistent depth estimation. More visualization results for each task are presented in \textit{Appendix}.}  
\label{fig:exp2} 
\vspace{-6mm}
\end{figure*}

\subsection{Experimental Setup}

\noindent \textbf{Training Dataset.} 
NutWorld is pre-trained on MiraData~\cite{ju2024miradatalargescalevideodataset} and RealEstate10K~\cite{zhou2018stereo}. MiraData is a high-quality video dataset consisting primarily of 3D engine-generated scenes and movie clips with diverse motion patterns. The RealEstate10K dataset contains indoor house tour videos that showcase various architectural scenes and camera movements.
~\footnote{Unlike previous Generalizable 3DGS approaches~\cite{chen2025mvsplat, pixelsplat, zhang2024transplat}, we treat RealEstate10K as a pure video dataset rather than a multi-view 3D scene dataset, thus not utilizing the COLMAP-calibrated camera poses}. 
During pre-processing, we segment the original videos into video cubes, each containing 10 consecutive frames as the basic processing unit. Detailed information on the description of the dataset, preprocessing, and splitting of trains is provided in the \textit{Appendix}.





\noindent\textbf{Implementation Details.} NutWorld is trained on 32 NVIDIA A100 (80GB) GPUs with a batch size of 256 for around 4 days. 
To improve computational efficiency, we integrate Flash-Attention-v2~\cite{dao2022flashattention, dao2023flashattention}, gradient checkpointing~\cite{sohoni2019low}, and mixed-precision training with BF16~\cite{zamirai2020revisiting}.
The orthographic camera coordinates are bounded to $[-1, 1]$ along the $x$ and $y$ axes and to $[0, 1]$ along the $z$ axis. The input frames are resized to $512 \times 288$ to preserve the aspect ratio.
We adopt a two-phase training strategy: a static phase using single frames ($K=1$) with window size $\mathcal{W} = 576$, followed by a dynamic phase where we initialize from the static weights and expand the window to $\mathcal{W} = 3456$ to allow spatio-temporal attention in the hierarchical upsampler. Additional training details, including hyper-parameters and network architecture, are provided in the \textit{Appendix}.

\subsection{Video Reconstruction}


\noindent \textbf{Experimental Protocol.} We evaluated the video reconstruction performance of NutWorld on 50 randomly selected test video clips from RealEstate10K and MiraData, both with a default length of 90 frames via standard reconstruction quality metrics (PSNR, SSIM, and LPIPS~\cite{zhang2018perceptual}). As there are currently no other feed-forward dynamic Gaussian approaches, we compared with optimization-based methods including Splatter-a-Video (SaV)~\cite{sun2024splatter}, 4DGS~\cite{wu20244d}, RoDynRF~\cite{liu2023robust} and CoDeF~\cite{ouyang2024codef} as the most relevant baselines. For fair comparison, all methods incorporate the confined canonical space, depth and flow supervision. We used the official implementations for most methods while SaV was reproduced according to the implementation details provided in their paper. 

\noindent \textbf{Comparison with Baselines.}
We evaluate NutWorld's representation effectiveness through both qualitative and quantitative experiments on video reconstruction.
As shown in Figure~\ref{fig:main-comparison}, our pre-trained NutWorld effectively captures spatial details and temporal dynamics, outperforming both Gaussian-based SaV~\cite{sun2024splatter} and NeRF-based CoDeF~\cite{ouyang2024codef} in reconstruction quality. This superior performance can be attributed to STAG's elaborated deformable field and positional constraints, which provide more expressive and robust temporal modeling capabilities compared to SaV's Fourier series and CoDeF's 2D canonical representation.
Furthermore, as evidenced in Table~\ref{tab:main-comparison}, NutWorld achieves the best of both worlds in reconstruction quality and computational efficiency. Notably, 
NutWorld reconstructs a 90-frame video in just $1.8$ seconds, achieving a $1000\times$ speedup over optimization-based methods. Equipped with segment-based inference that limits the number of Gaussians per segment, NutWorld achieves a rendering speed of 450 FPS, substantially surpassing SaV's 149 FPS, which requires around $2 \times 10^6$ Gaussians for the same video.



\subsection{Video Downstream Tasks} 

Large-scale pretrained NutWorld empowers video applications including object segmentation, frame interpolation, video editing, novel view synthesis, and consistent depth prediction. We present representative qualitative results in Figure~\ref{fig:exp2}, with additional results provided in the \textit{Appendix}.

\noindent \textbf{Video object segmentation.} The explicit nature of STAG allows us to propagate object masks in a certain frame to subsequent frames. Specifically, the corresponding STAGs in the object mask can be explicitly selected in the first frame, following previous training free Gaussian segmentation methods~\cite{flashsplat, hu2024semantic}. As visualized in following Fig.~\ref{fig:point-track}, NutWorld predicts coherent deformation $u_i(t)$ for each Gaussian over time, object masks in subsequent frames can be rendered from the STAGs selected initially.
\begin{figure}[h!]
\vspace{-4mm}
\setlength{\abovecaptionskip}{0.1cm}
\setlength{\belowcaptionskip}{0cm}
\centering
\includegraphics[width=0.85\linewidth]{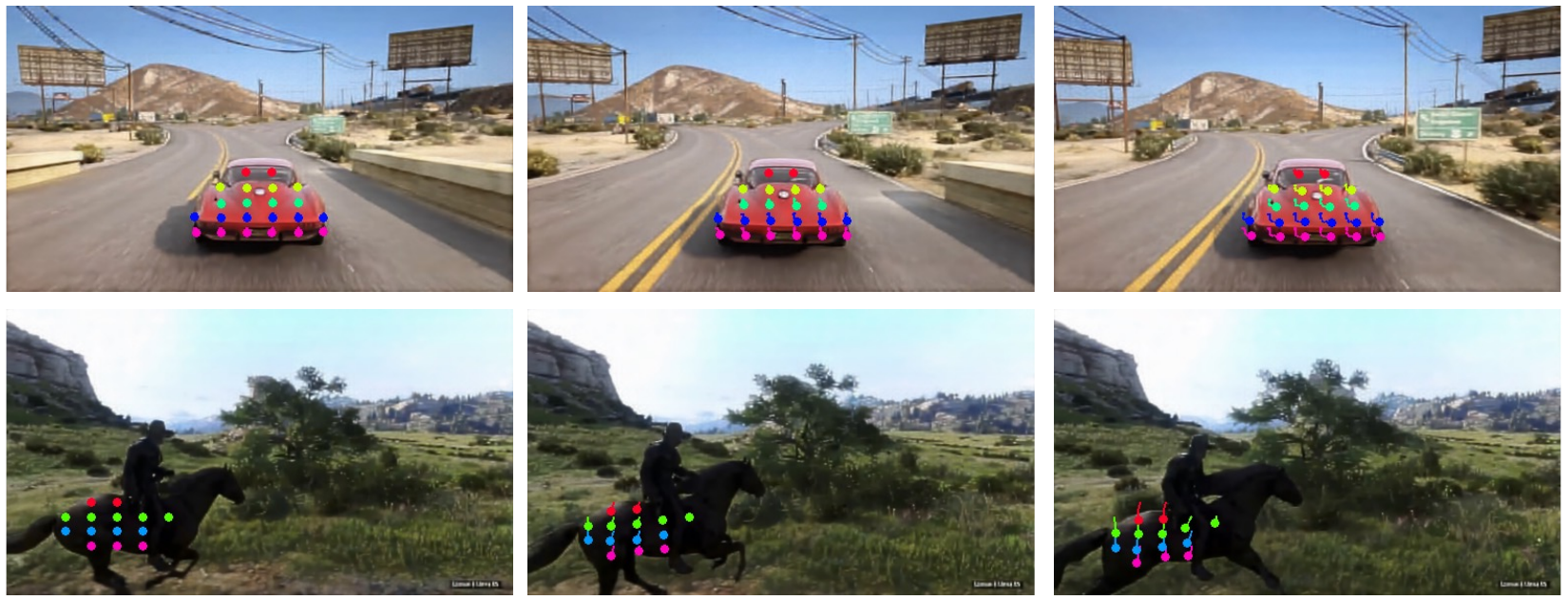}
\caption{Visualization of Gaussian trajectories. Trajectories of selected Gaussian centers are illustrated as point tracks.}
\vspace{-4mm}
\label{fig:point-track}
\end{figure}
In particular, this capability emerges as a by-product without specific training~\cite{sun2024splatter, wang2024gflow} in video Gaussian representation.

\noindent \textbf{Frame interpolation.} The continuous trajectories learned for STAGs, regularized by calibrated optical flow, enable temporal interpolation of scene dynamics at arbitrary FPS. These interpolated STAGs, with smoothly varying dynamic attributes, facilitate intermediate frame rendering, a capability beyond the scope of per-frame methods~\cite{ren2024l4gm}.

\noindent \textbf{Consistent depth prediction.} The calibrated depth regularization prevents depth collapse while maintaining temporally coherent spatial configurations in scene geometry. Additionally, NutWorld demonstrates potential for distilling other image features, such as SAM~\cite{kirillov2023segment} and CLIP~\cite{radford2021learning}, which we consider a promising direction for future work.

\noindent \textbf{Video editing.} By integrating with a MLLM-guided editing model~\cite{fu2023guiding}, NutWorld enables precise frame-level painting and stylization by optimizing the sliced STAG representation. These edits propagate temporally while maintaining visual coherence throughout the video sequence. The visualization results are provided in the \textit{Appendix}.

\noindent \textbf{Novel view synthesis.} NutWorld achieves novel view synthesis within practical bounds by incorporating depth priors to mitigate spatial ambiguity. Camera extrinsic adjustment enables novel view rendering, while camera intrinsic manipulation allows for effects such as dolly zoom. Please refers to \textit{Appendix} for the visualization results.



\section{Ablation Study} 



We analyze NutWorld's design choices through ablation studies on the 50 selected video clips. As shown in Table~\ref{tab:ablation}, our experiments demonstrate that discarding any component from the multi-component pipeline leads to significant performance degradation.


\begin{table}[h!]
\setlength{\abovecaptionskip}{0.1cm}
\setlength{\belowcaptionskip}{0cm}
\vspace{-4mm}
  \centering
  \setlength{\tabcolsep}{8pt}  
  \renewcommand{\arraystretch}{1.2}  
  \resizebox{0.75\columnwidth}{!}{\begin{tabular}{l S[table-format=2.2] S[table-format=1.4] S[table-format=1.4]}
    \toprule
    \textbf{Method} & {\textbf{PSNR}$\uparrow$} & {\textbf{SSIM}$\uparrow$} & {\textbf{LPIPS}$\downarrow$} \\
    \midrule
    w/o Flow Loss & {26.39} & {0.8345} & {0.2482} \\
    w/o Depth Loss & {28.15} & {0.8751} & {0.1858} \\
    w/o STAG & {19.58} & {0.6255} & {0.4713} \\
    Ours ($n = 2$) & {29.18} & {0.9015} & {0.1415} \\
    Ours ($n = 3$) & {31.15} & {0.9266} & {0.1385} \\
    \bottomrule
  \end{tabular}}
  \caption{Ablation study on component-wise contribution. $n$ represents the number of upsampler blocks.} 
  \label{tab:ablation}
  \vspace{-4mm}
  \end{table}

\noindent \textbf{Ablations on STAG representation.}
To validate the effectiveness of STAG representation, we perform an ablation by loosening its positional constraints. Following~\cite{tang2025lgm}, we implement a less constrained variant where Gaussian positions are predicted with only $tanh$ activation, limiting their range to $[-1, 1]$. As shown in Table~\ref{tab:ablation}, this loosened constraint leads to significantly degraded performance by \underline{$10dB$} decrease in PSNR, with slower convergence, blurred artifacts and unstable optimization behavior.
These results demonstrate the necessity of structured positional constraints, as the unconstrained Gaussians introduce additional spatial ambiguity during alpha compositing. In contrast, STAG's localized positional constraints provide explicit spatial and temporal correspondence, enabling efficient optimization and high-quality rendering.

\begin{figure}[h!]
\vspace{-4mm}
\setlength{\abovecaptionskip}{0.1cm}
\setlength{\belowcaptionskip}{0cm}
\centering
\includegraphics[width=1.0\columnwidth]{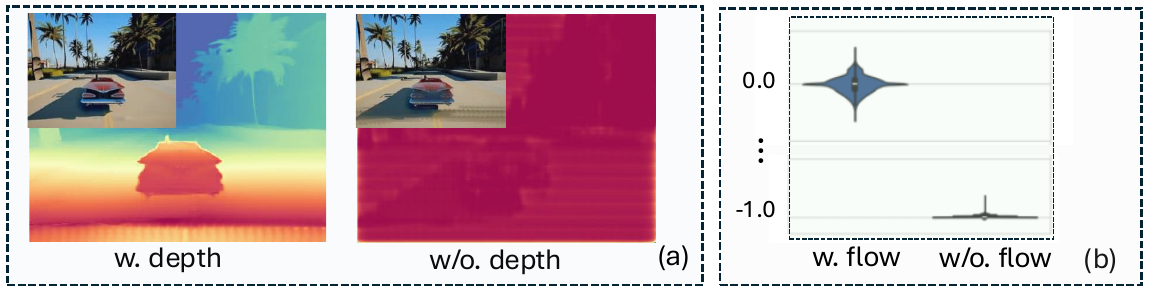}
\caption{Qualitative ablation on 2D prior regularization.}
\label{fig:prior_ablation}
\vspace{-4mm}
\end{figure}

\noindent\textbf{Ablations on depth prior}. To evaluate the depth prior (Eq.~\ref{eq:depth_prior}), we trained the NutWorld variant without depth supervision in Figure~\ref{fig:prior_ablation}(a) for comparison, which reveals that the variant w/o depth tend to lost spatial arrangement and converges to a collapsed shortcut instead, i.e. all STAGs are splattered onto a single $z$ plane. Furthermore, quantitative experiments in Table~\ref{tab:ablation} reveal that removing the depth prior leads to a degraded rendering quality, as evidenced by the decrease in PSNR from $29.18 dB$ to $28.15 dB$. These results highlight the necessity of depth prior to address the spatial ambiguity in NutWorld.

\noindent\textbf{Ablations on flow prior}. To evaluate the flow prior (Eq.~\ref{eq:flow_prior}), we trained a NutWorld variant without flow supervision for comparison. The distribution of the deformation field $\mu(t)$ across $K=6$ frames is visualized in Figure~\ref{fig:prior_ablation}(b) via a violin plot. Without flow supervision, the model exhibits large deformation values with low variance, causing STAGs to deviate from the canonical space in non-reference frames as defined in Eq.~\ref{eq:temporal_align}. This indicates that the variant w/o flow tends to learn an undesirable shortcut by representing each frame with independent STAGs, leading to temporal discontinuity. In contrast, with flow supervision, the distributions are centered near zero with appropriate variance, demonstrating that NutWorld could recover temporal motion through the flow prior, which effectively prevents such shortcut behavior. Additionally, quantitative experiments in Table~\ref{tab:ablation} show that temporal discontinuity leads to inferior reconstruction quality, especially for complex motions.



\section{Conclusion}
In this paper, we introduce NutWorld, a novel framework for efficiently representing casual monocular videos through dynamic Gaussian Splatting. By introducing the structured STAG representation and incorporating effective depth and flow regularization, our approach successfully tackles several fundamental challenges in monocular video representation, achieving both spatial and temporal coherence without per-scene optimization. Comprehensive experiments demonstrate that NutWorld not only achieves high-fidelity video reconstruction in real-time but also enables various downstream applications. In the future, distilling rich visual features (e.g., SAM, CLIP) into our STAG representation and adapting our representation paradigm for video generation tasks are promising directions to explore.

\clearpage

\maketitlesupplementary
\setcounter{page}{1}
\setcounter{section}{0}

\noindent The \textit{Appendix} is organized as follows:
\begin{itemize}[leftmargin=*]
    \item \textbf{Appendix~\ref{app:imp}:} provides additional details about the Nutworld pipeline, including the implementation of orthographic rasterization, Gaussian decoder, flow prior calibration, and segment-based inference.
    
    \item \textbf{Appendix~\ref{app:model}:} provides further details on experiment design and model configuration.
    

    \item \textbf{Appendix~\ref{app:result}:} presents additional experimental results on video reconstruction and downstream tasks.

        \item \textbf{Appendix~\ref{app:limit}:} discuss the limitations and potential future directions of NutWorld.

\end{itemize}

\section{Implementation Details}
\label{app:imp}

\noindent\textbf{Orthographic rasterization.}
Orthographic projection is leveraged to circumvent the necessity for explicit camera pose estimation in our setting. 
Specifically, we employ a fixed orthographic camera model and modify the EWA projection~\cite{zwicker2002ewa} from perspective to orthographic in Gaussian rasterization. The EWA projection in original rasterization is formulated as:
\begin{equation}
\Sigma' = J W \Sigma W^T J^T,
\end{equation}
where \(J\) represents the Jacobian matrix of the projective transformation. In the case of perspective projection, the Jacobian \(J\) is formulated as:
\begin{equation}
\begin{aligned}
(u, v) &= \left( f_x \cdot x / z + c_x, f_y \cdot y / z + c_y \right), \\
J = \frac{\partial(u, v)}{\partial(x, y, z)} &= 
\begin{pmatrix}
f_x / z & 0 & -f_x \cdot x / z^2 \\
0 & f_y / z & -f_y \cdot y / z^2
\end{pmatrix}.
\end{aligned}
\end{equation}
In contrast, for orthographic projection, the EWA projection is modified as follows:
\begin{equation}
\begin{aligned}
(u, v) &= \left( f_x \cdot x + c_x, f_y \cdot y + c_y \right), \\
J = &\frac{\partial(u, v)}{\partial(x, y, z)} = 
\begin{pmatrix}
f_x & 0 & 0 \\
0 & f_y & 0
\end{pmatrix}.
\end{aligned}
\end{equation}
In this formulation, the near and far planes are set to $0$ and $1$, respectively, constraining the $z$ axis within this range. Additionally, the $x$ and $y$ axes are constrained to $[-1, 1]$ to facilitate structured prediction.


\noindent\textbf{STAG Parameterization.}
The parameterization of output parameters significantly impacts the model's convergence, despite STAG provides a relatively structured Gaussian representation. For reproducibility, we provide detailed configuration of the STAG decoder parameterization in Table~\ref{tab:act}. Common activation functions such as $\text{sigmoid}$, $\text{softplus}$, and $\text{normalize}$ are employed for most static Gaussian attributes, following previous works~\cite{pixelsplat, xu2024grm, tang2025lgm}. 
For the spatial aligned 3D Gaussian position, we predict it as $\mu = (u + \Delta_x, v + \Delta_y, d)$, where $u$ and $v$ are the aligned 2D pixel positions within the orthographic space, and both $\Delta_x$ and $\Delta_y$ are constrained using the $\tanh$ activation. For the depth value $d$, the $z$ axis in $[0, 1]$ is divided into 20 discrete bins. A discrete distribution is predicted over these bins, and the expectation is computed to robustly predict the depth value $d$.
For the deformable fields $\mu(t)$, the query timestamp $t$ is represented as $L=10$ sinusoidal expansions. The output of $\mathcal{F}_{\theta}$ remains unbounded, allowing invisible Gaussians to be driven out of the view space as needed.

\begin{table}[h]
\centering
\renewcommand{\arraystretch}{1.3}
\begin{tabular}{l|l}
\toprule
\rowcolor[HTML]{EFEFEF} \textbf{Attribute} & \textbf{parameterization} \\
\midrule
Position (static) & $\begin{array}{l}
x,y: \text{map} + \text{max} \cdot \tanh(\cdot) \\

\end{array}$ \\
\midrule
Position (dynamic) & unbounded \\
\midrule 
Depth & bin regression with \text{softmax} \\
\midrule
Scale & $0.1 \cdot \text{softplus}$ \\
\midrule
Rotation  & $\text{normalize}$ \\
\midrule
Opacity & $\text{sigmoid}$ \\
\midrule
RGB & $\text{sigmoid}$ \\
\midrule
\bottomrule
\end{tabular}
\caption{Detailed STAG parameterization.}
\label{tab:act}
\end{table}

\noindent\textbf{Flow Prior Calibration.} The estimated global optical flow between video frames often contains noise, which can hinder model convergence during training. To mitigate these noises, we employ a calibration strategy for the flow regularization loss $\mathcal{L}_{\text{flow}}$. Specifically, when a STAG moves out of the view space, we set the corresponding flow mask value $M_{i, k} = 0$. Additionally, we filter out flows with the top $20\%$ motion magnitudes as outliers. With these adjustments, the calibrated flow regularization loss is expressed as:
\begin{equation}
\mathcal{L}_{\text{flow}} = \sum_{i=1}^{K \times U \times V}\sum_{k=0}^{K-1} M_{i, k}\|\bar{\mu}_{i}(t_k) - \mu^*_i(t_k)\|_1,
\label{eq:flow_cal}
\end{equation}
This calibration ensures that only reliable motion information contributes to the training process, reducing the impact of noise and extreme outliers, encouraging the model to learn coherent motion patterns effectively.

\noindent\textbf{Segment-based inference.} 
We adopt a segment-based inference approach to transform long casual videos into STAG representations. This strategy processes video sequences using a sliding window mechanism, where each window contains one overlapping frame. Temporal coherence between segments is maintained through spatially aligned Gaussians in the overlapping frames shared by adjacent segments. The coherence is achieved through token-wise correspondence, as spatial positions in STAG directly correspond to identical pixel locations.
Through quantitative comparison shown in Table.1 in the manuscript, we further demonstrate that this segment-based inference strategy not only efficiently processes video segments in parallel but also maintains a manageable number of Gaussians for each segment due to the slicing window design. In contrast, SaV~\cite{sun2024splatter} directly uses millions of Gaussians to represent entire frames, resulting in significantly longer rendering times per frame.

\section{Experiment Configuration}
\label{app:model}
\noindent\textbf{Network Configuration}. As illustrated in Table~\ref{tab:conifg}, the transformer-based encoder processes input frames at $288 \times 512$ resolution, with concatenated RGB and depth channels. The architecture comprises 10 attention layers with 12 heads and 768-dimensional features, operating on $16 \times 16$ patches. The hierarchical upsampler incorporates 3 blocks, each containing 2 attention layers with a window size of 3456. Through these blocks, the channel dimension progressively decreases from 768 to 64, while spatial resolution doubles at each stage. With $n=2$ upsampler blocks, our model processes $K=6$ input frames to produce feature maps of spatial resolution $(U, V) = (128, 72)$, generating $55,296$ dynamic Gaussians. The STAG decoder implements a lightweight MLP with a single 1024-dimensional hidden layer, utilizing attribute-specific activation functions: linear for position, $\tanh$ for dynamics, $Softplus$ for scale, normalization for rotation, and $Sigmoid$ for opacity and color.




\noindent\textbf{Dataset Settings.} NutWorld is pre-trained on MiraData~\cite{ju2024miradatalargescalevideodataset} and RealEstate10K~\cite{zhou2018stereo}. For MiraData, we utilize its middle version containing approximately 34K video clips, each with a duration of about 2 minutes. To ensure data quality, we performed rigorous filtering by removing clips containing gaming pop-ups, black screens, and other artifacts, resulting in a curated dataset of 22K video clips~\cite{yi2023invariant}. The dataset is split randomly with a 95:5 ratio for training and testing. For RealEstate10K, unlike previous Generalizable 3DGS approaches~\cite{chen2025mvsplat, pixelsplat, zhang2024transplat}, we treat it as a pure video dataset rather than a multi-view 3D scene dataset, without utilizing COLMAP-calibrated camera poses. This dataset is similarly split with a 95:5 training-testing ratio. For evaluation, we randomly selected 40 videos from the MiraData test set and 10 from the RealEstate10K test set for both qualitative and quantitative comparison.
\begin{table*}[t]
\centering
\caption{Model Configuration of NutWorld}
\label{tab:modelconfig}
\vspace{-6pt}
\renewcommand{\arraystretch}{1.2}%
\begin{tabular}{>{\bfseries}p{0.4\linewidth} p{0.5\linewidth}}
\toprule
\rowcolor[HTML]{EFEFEF} 
\textbf{Parameter} & \textbf{Value} \\
\midrule
\multicolumn{2}{l}{\textbf{Encoder}} \\
\quad Input Resolution & $288 \times 512$ \\
\quad Input Channels & RGB (3) + Depth (1) \\
\quad Patch Size & $16$ \\
\quad Hidden Dimension & $768$ \\
\quad Attention Layers & $10$ \\
\quad Attention Heads & $12$ \\
\quad Window Size & $3456$ \\
\quad Positional Embedding & None \\
\midrule
\multicolumn{2}{l}{\textbf{Upsampler Block}} \\
\quad Scale Factor & $2$ \\
\quad Decoder Ratio & $2.0$ / $3.0$ \\
\quad Channel Width Decay & $768 \rightarrow 64$ \\
\quad Attention Layers per Block & $2$ \\
\quad Window Size & $3456$ \\
\quad Number of Blocks & $4$ \\
\midrule
\multicolumn{2}{l}{\textbf{STAG Decoder}} \\

\quad MLP Dimension & $1024$ \\
\quad Hidden Layers & $1$ \\
\quad Position Activation & None \\
\quad Dynamic Activation & $tanh$ \\
\quad Scale Activation & $Softplus$ \\
\quad Rotation Activation & $Normalize$ \\
\quad Opacity Activation & $Sigmoid$ \\
\quad Color Activation & $Sigmoid$ \\

\midrule
\multicolumn{2}{l}{\textbf{Training Details}} \\
\quad Optimizer & AdamW \\
\quad Learning Rate & $1 \times 10^{-4}$ \\
\quad Weight Decay & $0.05$ \\
\quad Warm-Up Ratio & $0.04$ \\
\quad Batch Size & $4$  \\
\quad Number of GPUs & $32$ \\
\quad Total Epochs & $100$ \\
\quad Input Frames & $6$ \\
\quad Output Frames & $10$ \\
\quad Mixed Precision & bf16 \\
\midrule
\multicolumn{2}{l}{\textbf{Loss Weights}} \\
\quad $\lambda_{\text{MSE}}$ & $0.5$  \\
\quad $\lambda_{\text{depth}}$ & $0.001$  \\
\quad $\lambda_{\text{flow}}$ & $0.2$ \\
\bottomrule
\end{tabular}
\label{tab:conifg}
\vspace{-12pt}
\end{table*}

\section{More Experiments Results}

Due to limited space in the manuscript, we provide more qualitative results on video reconstruction, object segmentation, frame interpolation, editing, novel view synthesis and depth prediction in the following figures. Note that we don't claim that NutWorld achieves state-of-the-art performance across all these downstream tasks. Instead, our focus is on demonstrating NutWorld as a versatile video representation framework with broad applicability and adaptability. We believe that with task-specific adaptations, NutWorld has the potential to compete with specialized state-of-the-art methods in these individual video domains. \textbf{Please refer to the attached material for video visualization.}

\begin{figure*}[t]
\centering 
\includegraphics[width=0.95\linewidth]{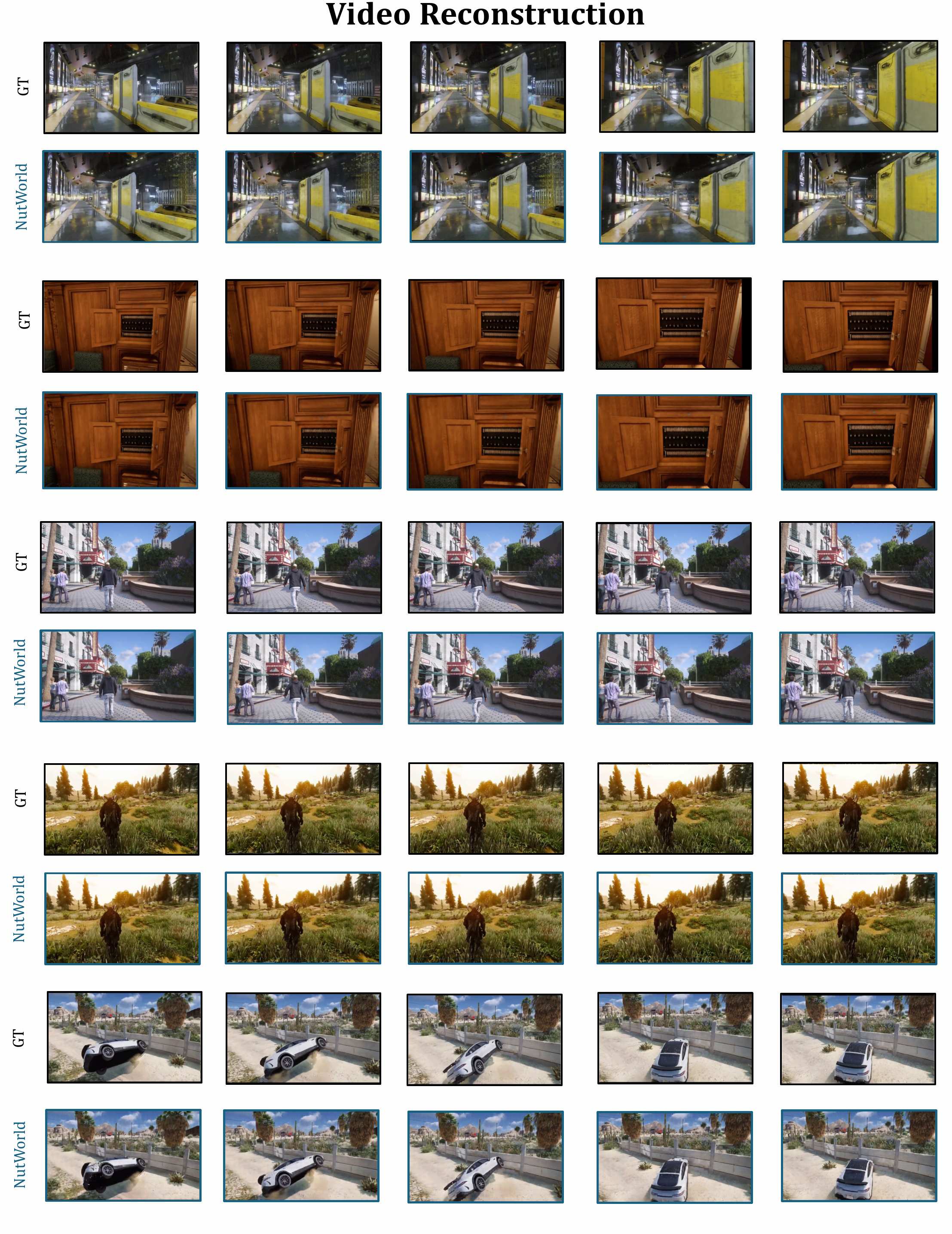} 

\caption{More qualitative results on video reconstruction.}  
\label{fig:app_re} 

\end{figure*}

\begin{figure*}[t]
\centering 
\includegraphics[width=0.98\linewidth]{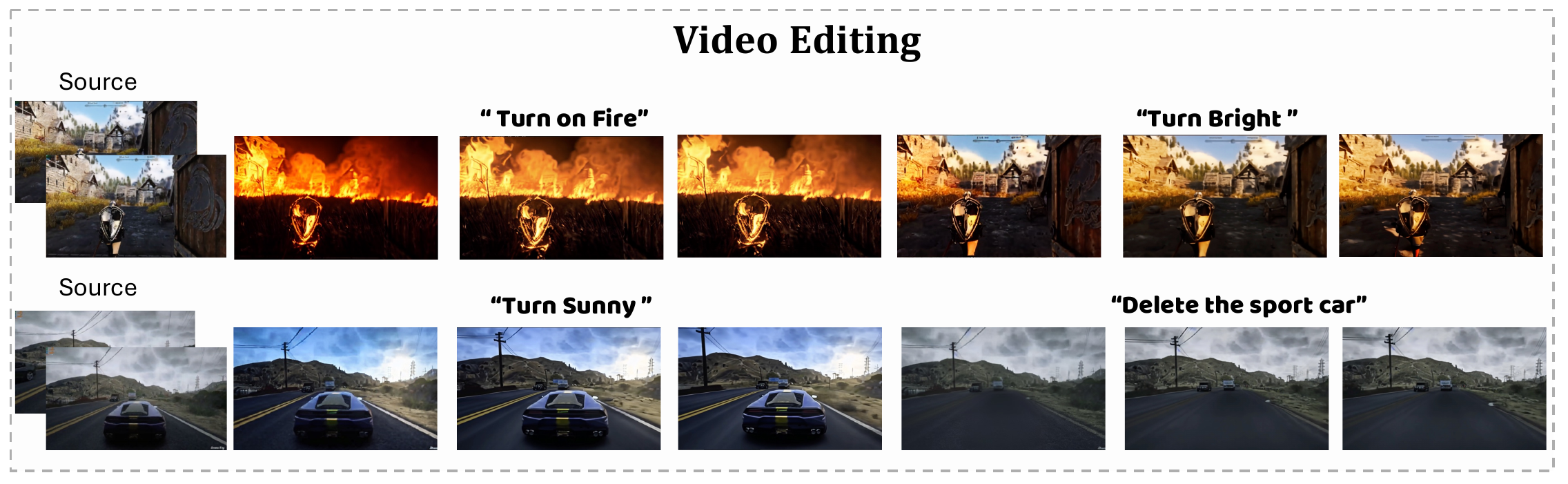} 

\caption{More qualitative results on video editing.}  
\label{fig:app_edit} 

\end{figure*}

\begin{figure*}[t]
\centering 
\includegraphics[width=0.96\linewidth]{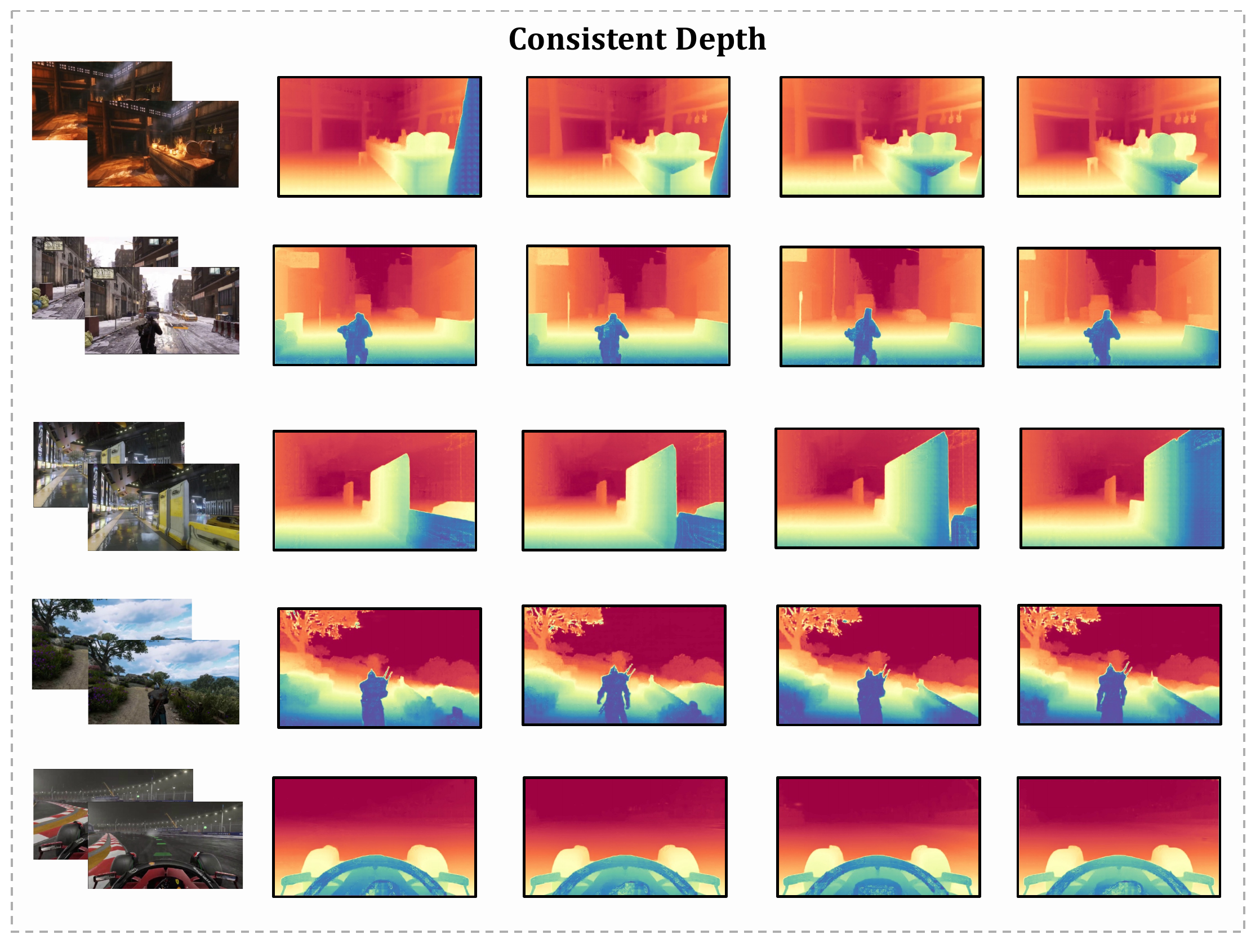} 

\caption{More qualitative results on consistent depth prediction.}  
\label{fig:app_depth} 

\end{figure*}

\begin{figure*}[t]
\centering 
\includegraphics[width=1.0\linewidth]{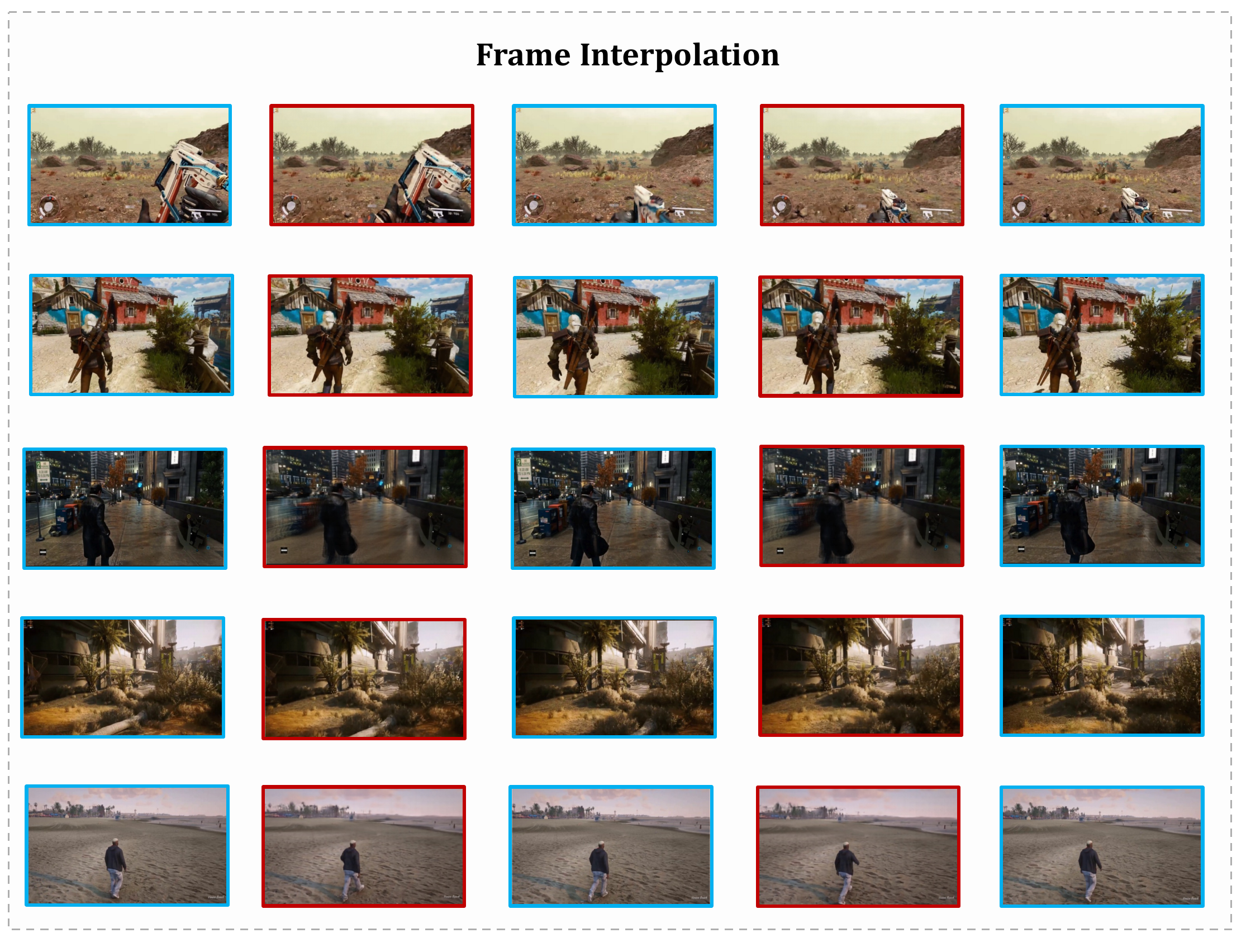} 

\caption{More qualitative results on frame interpolation. Note that \textbf{\textcolor{red}{Red Frame}} denotes the interpolated frame.}  
\label{fig:app_frame} 

\end{figure*}

\begin{figure*}[t]
\centering 
\includegraphics[width=1.0\linewidth]{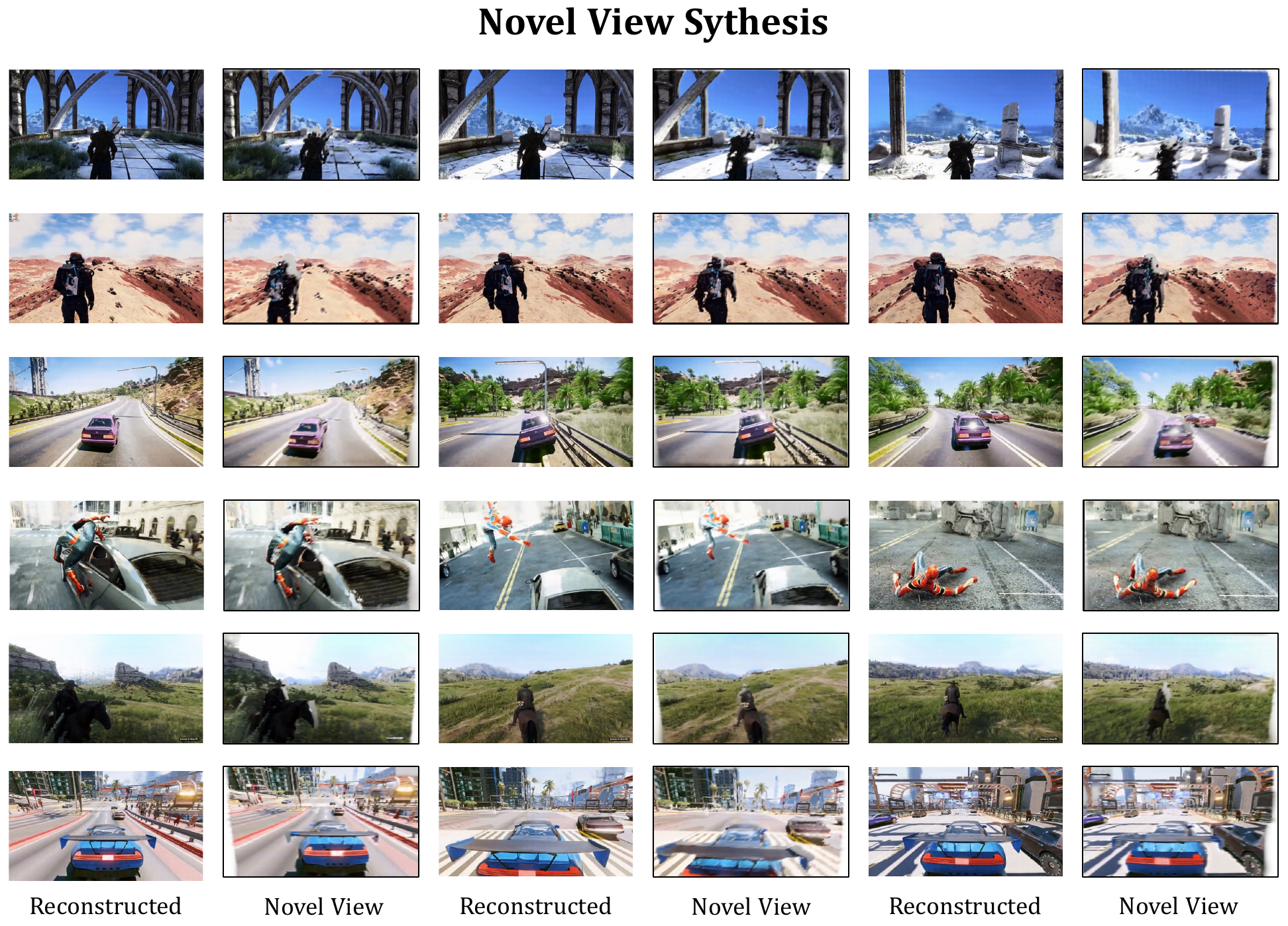} 

\caption{Qualitative results on novel view synthesis.}
\label{fig:app_vis} 

\end{figure*}

\begin{figure*}[b]
\centering 
\includegraphics[width=1.0\linewidth]{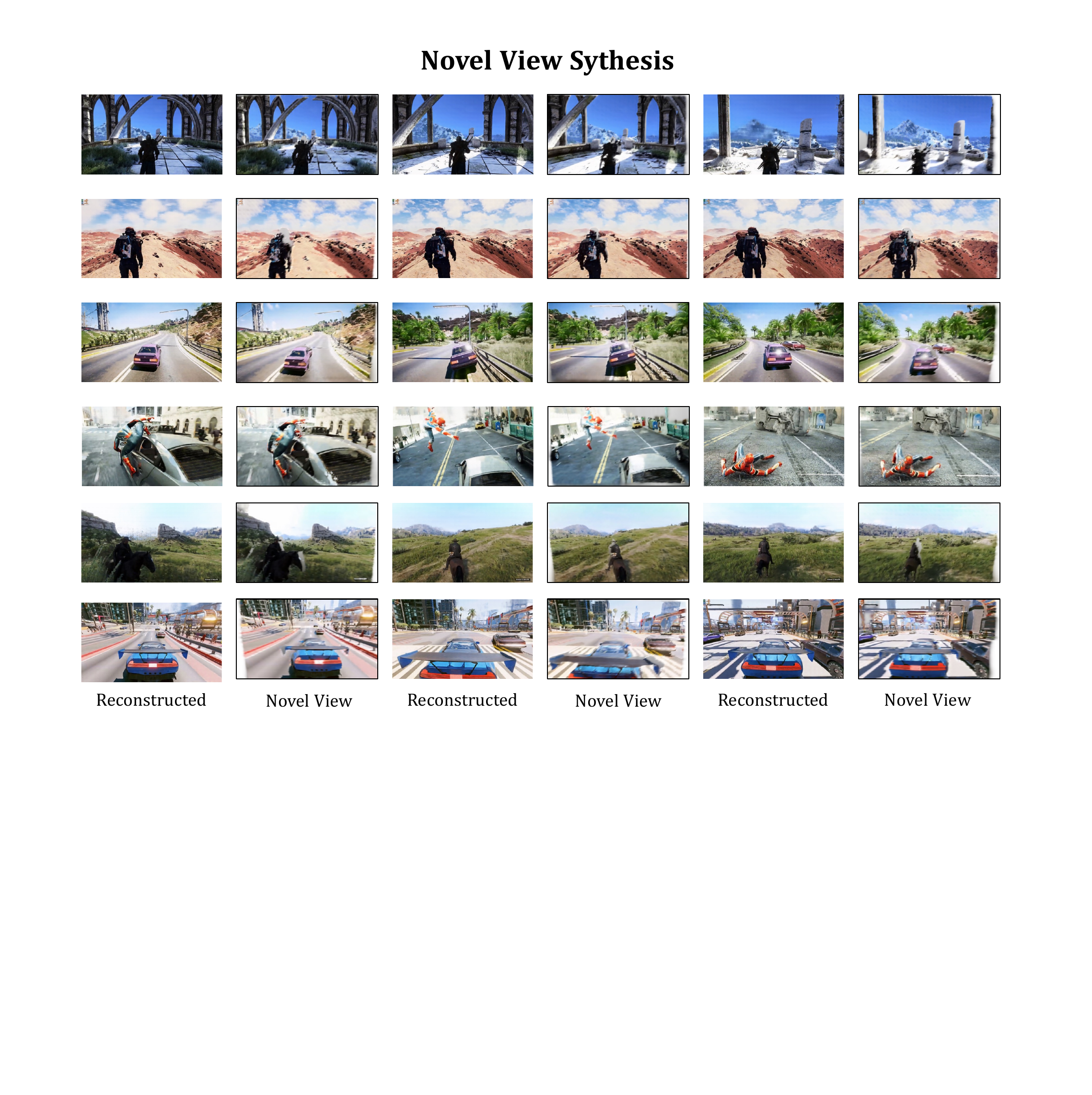} 

\caption{Qualitative results on video object segmentation.}
\label{fig:app_vos} 
\end{figure*}

\label{app:result}

\section{Limitations}
\label{app:limit}
While our NutWorld framework demonstrates significant advances in spatial-temporal video modeling and rendering efficiency, several limitations warrant discussion. (1) The framework's reliance on depth and optical flow priors makes its performance inherently dependent on external models, which may not generalize effectively to challenging scenarios involving complex motion or suboptimal lighting conditions. (2) While the segment-based inference strategy enables accelerated rendering, it introduces a trade-off between processing speed and global temporal consistency by prioritizing localized frame segments over the complete video sequence. (3) Despite achieving substantial runtime efficiency, the framework's training process remains computationally intensive, potentially limiting its deployment on resource-constrained edge devices. (4) Unlike Langsplat~\cite{qin2024langsplat} and latentsplat~\cite{wewer2024latentsplat}, our current framework does not embed latent or semantic features into each STAG, necessitating integration with other pre-trained models during inference for high-level vision tasks such as editing and reasoning. However, future work could explore distilling rich visual features (e.g., SAM, CLIP) into our STAG representation and adapting our representation paradigm for video reasoning and generation tasks.

\clearpage 
\newpage












\clearpage
\newpage


\end{document}